\documentclass[conference]{IEEEtran}
\usepackage{times}

\usepackage[numbers]{natbib}
\usepackage{multicol}
\usepackage[bookmarks=true]{hyperref}
\usepackage{graphicx}
\usepackage{xspace}
\usepackage{amsmath}
\usepackage{amssymb}
\usepackage[utf8]{inputenc} %
\usepackage[T1]{fontenc}    %
\usepackage{amsfonts}       %
\usepackage{nicefrac}       %
\usepackage{microtype}      %

\usepackage{algorithm}
\usepackage{algpseudocode}
\usepackage{times}
\usepackage{graphicx}
\usepackage{caption}
\usepackage{listings}[language=Python]
\usepackage{color}
\usepackage[table,xcdraw,dvipsnames,usenames]{xcolor}
\hypersetup{
	colorlinks=true,
	linkcolor=orange,
	filecolor=magenta,      
	urlcolor=orange,
	citecolor=orange,
}
\usepackage{booktabs}
\usepackage{multirow}
\usepackage{natbib}
\usepackage[capitalise, nameinlink]{cleveref}
\newcounter{savefootnote}
\newcounter{symfootnote}
\newcommand{\symfootnote}[1]{%
   \setcounter{savefootnote}{\value{footnote}}%
   \setcounter{footnote}{\value{symfootnote}}%
   \ifnum\value{footnote}>8\setcounter{footnote}{0}\fi%
   \let\oldthefootnote=\thefootnote%
   \renewcommand{\thefootnote}{\fnsymbol{footnote}}%
   \footnote{#1}%
   \let\thefootnote=\oldthefootnote%
   \setcounter{symfootnote}{\value{footnote}}%
   \setcounter{footnote}{\value{savefootnote}}%
}
\newcommand{\website}{\url{https://mimic-video.github.io}}

\newcommand{\model}{mimic-video\xspace}
\begin{document}
\title{mimic-video: Video-Action Models for Generalizable Robot Control Beyond VLAs}
\author{Jonas Pai$^{* 1,2,3}$, Liam Achenbach$^{* 1,2,3}$, Victoriano Montesinos$^{1}$, Benedek Forrai$^{1}$, Oier Mees$^{\dagger 2,5}$, Elvis Nava$^{\dagger 1,3,4}$ 
\\
${}^{1}$mimic robotics ${}^{2}$Microsoft Zurich ${}^{3}$ETH Zurich ${}^{4}$ETH AI Center ${}^{5}$UC Berkeley\\
{\small $^*$Core Contributors $^\dagger$ Co-advising}\\
\website}

\makeatletter
\let\@oldmaketitle\@maketitle%
\renewcommand{\@maketitle}{\@oldmaketitle
  \begin{center}
  \captionsetup{type=figure}
  \includegraphics[width=\textwidth]{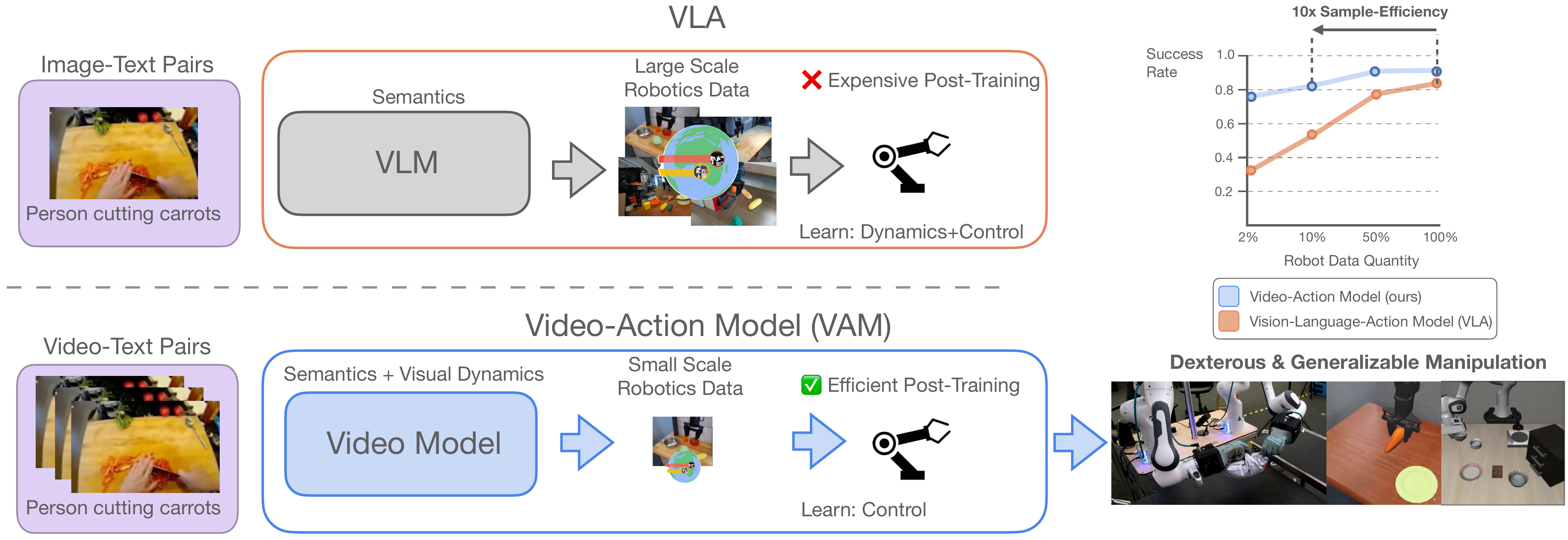}
    \captionof{figure}{\small We introduce \model, a new class of Video-Action Model (VAM) that grounds robotic policies in pretrained video models. Unlike standard VLAs that must learn physical dynamics from scratch (top), \model leverages the inherent visual dynamics of video backbones to isolate the control problem (bottom). This enables state-of-the-art performance on dexterous manipulation tasks, while achieving 10x greater sample efficiency compared to VLAs (right).} 
    \vspace{-1.5em}
    \label{fig:teaser}
    \end{center}
}
\makeatother

\maketitle
\addtocounter{figure}{-1}

\maketitle

\begin{abstract}
Prevailing Vision-Language-Action Models (VLAs) for robotic manipulation are built upon vision-language backbones pretrained on large-scale, but disconnected static web data. As a result, despite improved semantic generalization, the policy must implicitly infer complex physical dynamics and temporal dependencies solely from robot trajectories. This reliance creates an unsustainable data burden, necessitating continuous, large-scale expert data collection to compensate for the lack of innate physical understanding. We contend that while vision-language pretraining effectively captures semantic priors, it remains blind to physical causality. A more effective paradigm leverages video to jointly capture semantics and visual dynamics during pretraining, thereby isolating the remaining task of low-level control. To this end, we introduce \model, a novel Video-Action Model (VAM) that pairs a pretrained Internet-scale video model with a flow matching-based action decoder conditioned on its latent representations. The decoder serves as an Inverse Dynamics Model (IDM), generating low-level robot actions from the latent representation of video-space action plans. Our extensive evaluation shows that our approach achieves state-of-the-art performance on simulated and real-world robotic manipulation tasks, improving sample efficiency by 10x and convergence speed by 2x compared to traditional VLA architectures.
\end{abstract}

\IEEEpeerreviewmaketitle

\section{Introduction}

Building on the capabilities of pretrained Vision-Language Models (VLMs), Vision-Language-Action Models (VLAs) transfer semantic knowledge acquired through Internet-scale vision-language pretraining to the physical domain. By finetuning a VLM on diverse robot action data, VLAs learn generalist, natural language-conditioned robot manipulation policies that combine a wide range of skills and exhibit impressive generalization to unseen instructions, objects, and environments \cite{zitkovich_rt-2_2023, kim_openvla_2024, black_pi0_2024, intelligence__05_2025}.

However, this paradigm faces a fundamental limitation: the pretraining data, while massive in scale, is inherently static. Images and text lack explicit, temporally-grounded information about dynamics and physical procedures that are crucial for complex manipulation. Consequently, the burden of learning physical dynamics (how objects move, deform and interact) falls entirely on the post-training stage, where the model must infer these from scarce and expensive expert-teleoperated demonstrations. This heavy reliance on robot data creates a data-efficiency bottleneck that limits scalability. 
While prior works have explored augmenting VLA training with auxiliary video-derived signals such as language plans, affordances, or keypoints~\cite{zawalski_robotic_2025,gao2025radtrainingendtoenddriving,huang2025thinkactvisionlanguageactionreasoningreinforced,intelligence__05_2025}, reducing dense video into such sparse representations creates an information bottleneck, failing to capture fine-grained dynamics.

In this work, we posit that the key to more sample-efficient and capable robot policies lies in leveraging a pretraining modality that inherently encodes dynamic, procedural information: video. Unlike static image-text pairs, internet-scale video data provides rich knowledge on ``how things are done'', capturing the nuanced physics of interaction, how objects move, deform and react to forces. However, effectively harnessing this data remains a significant challenge. Prior approaches leveraging pretrained video models typically learn the joint distribution over video and actions, factorized such that the predicted actions are conditioned on synthesized future frames \cite{liang_video_2025, liang_dreamitate_2024,du_video_2023}. However, recovering the policy typically requires fully generating these future frames, necessitating prohibitive video synthesis at every control step.

To address these limitations, we propose a more direct paradigm: grounding robot policies directly in the latent representations of a generative, pretrained video model. We introduce \model, a novel Video-Action Model (VAM) that unifies video modeling with robot control. Built upon a state-of-the-art video diffusion backbone, \model functions by first synthesizing a visual plan: given an initial observation and language instruction, the video backbone predicts a future trajectory within a compact latent space.
Rather than requiring full or even partial video generation, we extract intermediate video model representations to condition a downstream action decoder. This decoder operates as an Inverse Dynamics Model (IDM) to recover low-level motor commands.
This formulation allows the video backbone to remain frozen, eliminating the need to train it on scarce robot action data. Fundamentally, this architecture decouples the inherent multi-modality of long-horizon planning, now offloaded to the video backbone, from the downstream control task. This effectively frees the action decoder from modeling complex future distributions, allowing it to dedicate its entire capacity to the far simpler, unimodal and non-causal problem of inverse dynamics~\cite{lynch2020learning,mees2022hulc}.

Our primary contribution is \model, a novel generalist robot policy that integrates generative video pretraining with flow matching-based control, establishing a new class of methods we term Video-Action Models (VAMs). We evaluate our approach across a diverse suite of robotic embodiments ranging from standard single-arm manipulation to bimanual dexterous tasks, demonstrating state-of-the-art results in both simulated benchmarks and challenging real-world environments. Our \model model achieves this performance while improving sample efficiency by 10x and convergence speed by 2x compared to traditional VLA architectures.

\section{Related Work}

\paragraph{Imitation Learning for Robot Control}
End-to-end imitation learning has become the dominant paradigm for training general-purpose robot manipulation policies, enabling robots to acquire complex skills directly from expert demonstrations. This approach, which maps raw sensory observations to actions, has benefited from advances in generative modeling in so far as models act as ``data sponges'', able to absorb large and diverse pretraining datasets~\cite{collaboration_open_2024, kim_openvla_2024, zitkovich_rt-2_2023, doshi_scaling_2024, reed_generalist_2022, bousmalis2023robocat} to achieve downstream generalization in action generation.

While early approaches like ACT \cite{zhao_learning_2023} used a VAE to model action chunks, the field has shifted toward iterative generative frameworks, popularized by Denoising Diffusion Probabilistic Models~\cite{ho_denoising_2020}. This class of methods, encompassing Diffusion Policy \cite{chi_diffusion_2024,dasari2024ingredients} and the Flow Matching~\cite{lipman_flow_2023} decoders of the $\pi_0$/$\pi_{0.5}$ series \cite{black_pi0_2024,intelligence__05_2025}, has become the state-of-the-art. These generative approaches excel at modeling multi-modal distributions of expert actions and form the technical foundation for modern robot imitation learning policies, including our own action decoder.

\paragraph{Vision-Language-Action (VLA) Models}
A major breakthrough in robot learning has been the paradigm of Vision-Language-Action (VLA) models, which are obtained by finetuning large, pretrained Vision-Language Models (VLMs) on robotics data. Models like RT-2 \cite{zitkovich_rt-2_2023}, OpenVLA \cite{kim_openvla_2024}, and the $\pi_0$/$\pi_{0.5}$ series \cite{black_pi0_2024, pertsch25-fast, intelligence__05_2025} leverage the vast semantic knowledge embedded in their backbones from pretraining on internet-scale image-text data. This allows them to follow open-ended language instructions, understand abstract concepts, and generalize to novel objects, environments, and tasks in a zero-shot fashion. However, a fundamental limitation of VLAs is that the VLM backbones they make use of are only pretrained with static vision and language data. They lack an inherent model of video dynamics, physics, or temporal progression, limiting their ability to reason about the physical consequences of actions. This critical knowledge must be learned from scratch from comparatively small and expensive robotics datasets.

Several works \cite{zawalski_robotic_2025, zhao_cot-vla_2025} make use of techniques like Chain-of-Thought reasoning~\cite{wei2022chain} in order to extract more useful grounded conditioning signals and representations~\cite{chen25training,huang2025thinkactvisionlanguageactionreasoningreinforced,zhao_cot-vla_2025} for VLAs. However, those approaches are still ultimately limited by relying on the static knowledge embedded in the pre-existing image-text pretraining of VLMs. They also typically result in significantly slower inference due to the computation of autoregressive plans before action decoding.

\paragraph{Video Models for Policy Learning}
The utilization of video prediction for robotic control has a long-standing history, primarily motivated by the potential to enable planning through visual foresight. Early works, such as the seminal approaches by \citet{oh_action-conditional_2015, watter_embed_2015, fragkiadaki_learning_2016, finn2016unsupervised, finn_deep_2017}, demonstrated how video prediction could enhance physical interaction. As generative models have matured to produce high-definition, coherent long-form content \cite{openai_video_2024, wiedemer_video_2025}, recent works have explored diverse integrations of video generation with policy learning.
The use of action-conditioned video models (world models) for policy learning has recently seen significant adoption. World models can help select more optimal action sequences at runtime by "imagining" their outcome \cite{assran_v-jepa_2025, qi_strengthening_2025}, or be used as learned simulators for evaluation and DAGGER-like \cite{ross2011reductionimitationlearningstructured} approaches \cite{guo2025ctrlworldcontrollablegenerativeworld,geminiroboticsteam2025evaluatinggeminiroboticspolicies}. In this work, we consider non action-conditioned video models.
One line of work fully generates pixel-space future video and obtains actions either via non-parametric methods such as tracking a custom end effector-mounted tool \cite{liang_dreamitate_2024}, or learned pixel-based Inverse Dynamics Models \cite{du2023learninguniversalpoliciestextguided, du_video_2023}. CoT-VLA \cite{zhao_cot-vla_2025} uses a pretrained VLM capable of generating images to generate a subgoal image and actions in one autoregressive sequence. Another line of work learns to model video during training without predicting video in action sampling; Unified World Models \cite{zhu_unified_2025} learns a model from scratch that can flexibly function as a policy, a video prediction model, or a forward or inverse dynamics model and LAPA \cite{ye_latent_2025} finetunes a VLM to predict ``latent actions'' (an encoding of the difference between the current and a future image), re-training only the output layer to predict actions in a subsequent training stage. FLARE \cite{zheng2025flarerobotlearningimplicit} aligns intermediate VLA representations with future vision-language embeddings, implicitly modeling video and actions jointly. Similarly to our approach, Video Policy \cite{liang_video_2025} explicitly models the joint video-action distribution and conditions a policy model on intermediate video model representations, but crucially does not allow for efficient sampling of the marginal action distribution.

Our proposed approach departs from most prior work by directly grounding control in the rich latent priors of internet-scale video models, rather than training from scratch or relying on pixel-level reconstruction. Additionally, by conditioning a lightweight inverse dynamics model on intermediate, noisy latent states, we bypass the computational cost of full video generation and the brittleness of heuristic tracking. This enables a scalable, end-to-end framework that effectively transfers broad physical understanding to downstream manipulation tasks, significantly reducing the reliance on expensive, large-scale robotic demonstrations.
\begin{figure}[t]
    \centering
\includegraphics[width=0.5\textwidth]{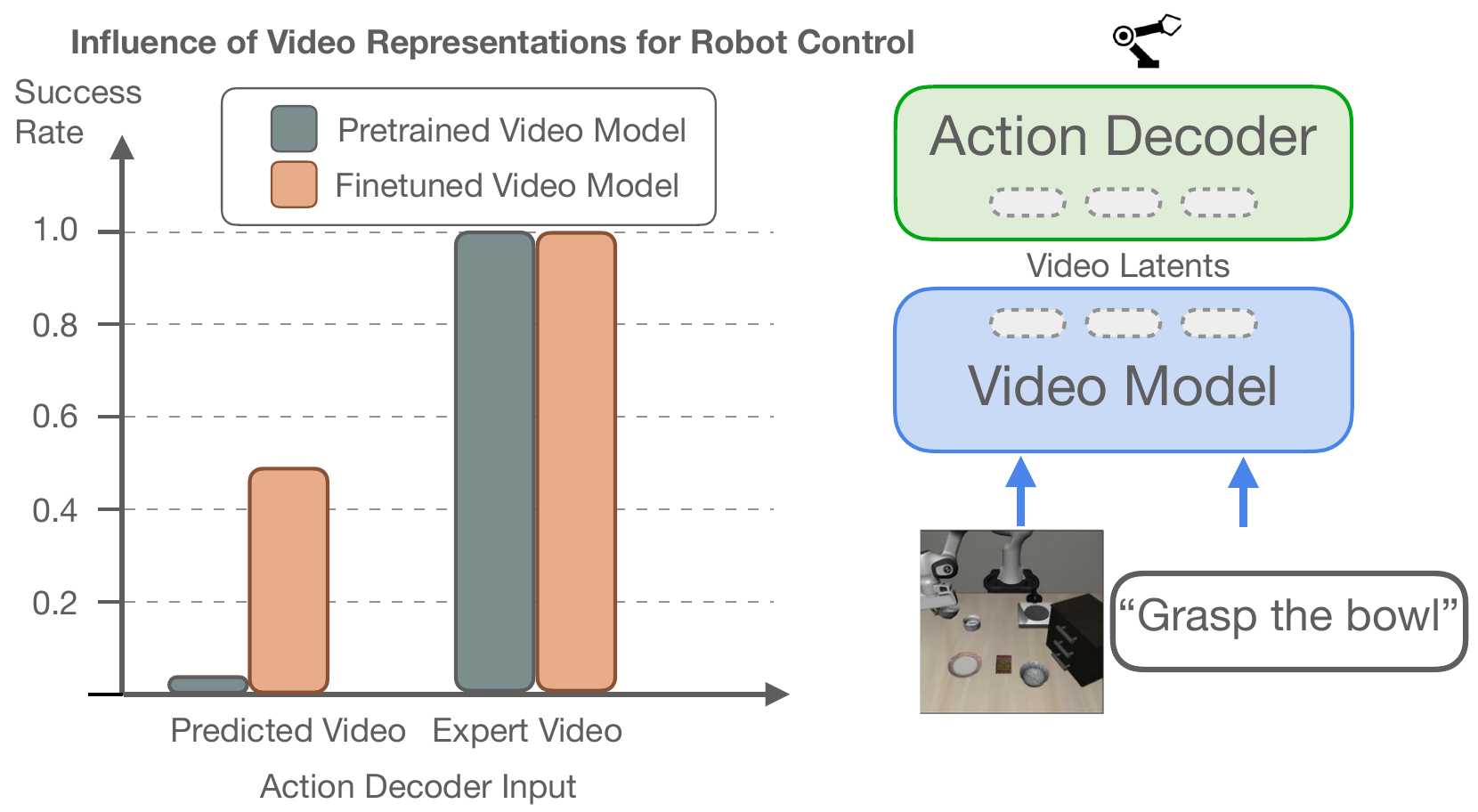}
    \caption{We compare success rates when conditioning our action decoder on different visual inputs: video latents generated by either predictions or ground-truth (expert) video for both features from a standard pretrained video model (gray), as well as a video model finetuned on video data from the robot dataset (orange). The near-perfect performance with ground truth inputs confirms that control effectively reduces to visual prediction, implying policy performance scales directly with video model quality.}
    \vspace{-0.5cm}
    \label{fig:perf-ours-gt}
\end{figure}

\section{Case Study: How does Video Generation Quality Affect Robot Policy Performance?}
\label{sec:case-study}

In this work, we argue that the internal representations arising in pretrained video model backbones are better suited for downstream robot learning compared to those in the VLM backbones commonly used in current state of the art VLAs. Intuitively, video models jointly model images, physical dynamics, alongside visual action plans. A policy trained on such representations effectively reduces the role of the action decoder to a simple translator, mapping visual action plans into a low-dimensional robot action trajectory. If this hypothesis holds, the bulk of learning in  Video-Action Models falls on the large-scale video pretraining and finetuning phases, while training the action decoder (the step requiring expensive, high-quality robot teleoperation data) becomes lightweight and data efficient.

We investigate this claim by conducting an ``oracle'' case study (see Fig.~\ref{fig:perf-ours-gt}), where we disentangle the difficulty of \textit{predicting} the future in robotic control tasks from \textit{executing} it. Concretely, we train an action decoder on top of video representations and evaluate its performance under different conditioning regimes.
We compare success rates when the decoder is conditioned on predicted video latents, from either a standard off-the-shelf video model or one finetuned on robotics data, versus ``oracle'' latents extracted from ground-truth future video frames. We observe a pronounced scaling behavior: while minimizing the domain gap via finetuning leads to improved performance when using \emph{predicted} video, conditioning on \emph{oracle} latents yields near perfect success rates regardless of whether the underlying backbone is finetuned on the target distribution or not. Notably, this finding suggests that  a high-quality pretrained video model backbone provides extremely rich representations for action decoding, sufficient on their own to perfectly decode low-level action plans with a decoder trained on minimal low-level action finetuning data. Consequently, the burden for policy learning in VAMs effectively shifts away from low-level action decoding towards video model pretraining and finetuning.

\section{Video-Action Models}
\label{sec:vam}

We introduce \model, a generative Video-Action Model (VAM) capable of modeling the joint distribution of video and robot actions. Our architecture couples two Conditional Flow Matching (CFM) models: a pretrained, language-conditioned video backbone and a lightweight action decoder that functions as an Inverse Dynamics Model (IDM) by conditioning on the video model's latent representations. 

\begin{figure*}[ht]
    \centering
    \includegraphics[width=0.9\textwidth]{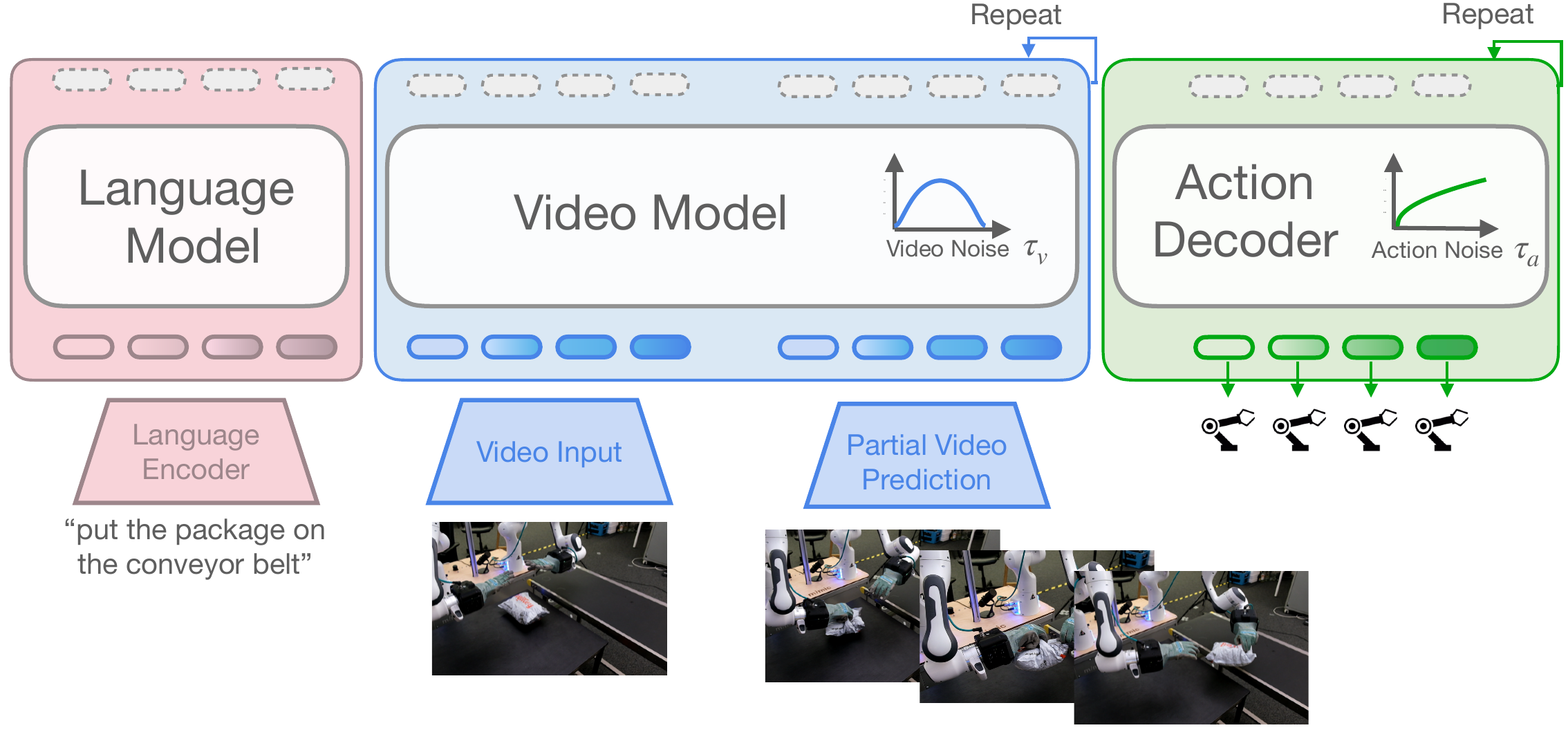}
    \caption{\model architecture: we instantiate our framework with a pretrained video generation backbone (Cosmos-Predict2~\cite{nvidia_world_2025,nvidia2025cosmosworldfoundationmodel}), which provides rich physical dynamics priors learned from large-scale video data. We adapt this model for control via a \textit{partial denoising} strategy, where the video backbone follows the flow to an intermediate flow time $\tau_v$ to extract latent visual plans. These representations condition a smaller action decoder, which processes proprioceptive states and predicts action trajectories. The video and action components operate on independent flow schedules ($\tau_v$ and $\tau_a$), allowing us to design the learning problem separately for each modality.}
    \label{fig:arch}
    \vspace{-0.5cm}
\end{figure*}

\subsection{Preliminaries: Flow Matching}
Both the video and action prediction components are trained using the Flow Matching framework~\cite{lipman_flow_2023} to model a data distribution $p_0(x^0)$ by constructing a Continuous Normalizing Flow \cite{chen2019neuralordinarydifferentialequations}. We use the conditional optimal transport path
\begin{equation}
\label{eq:x_tau}
    x^\tau = (1-\tau)x^0 + \tau \varepsilon,\quad \tau \in [0, 1]
\end{equation}
which interpolates between clean data $x^0$ (at $\tau=0$) and Gaussian noise $\varepsilon \sim \mathcal{N}(0, I)$ (at $\tau=1$) to define the conditional probability path $p_\tau\left(x^\tau \mid x^0\right)$.
The model parameterizes an estimator $v_\theta$ to the intractable marginal generating vector field
\begin{equation*}
    u_\tau(x^\tau) = \mathbb{E}_{p(x^0 \mid x^\tau)} u_\tau (x^\tau \mid x^0)
\end{equation*}
where $u_\tau(x^\tau \mid x^0) := \frac{d}{d\tau}x^\tau = \varepsilon - x^0$ is termed the conditional generating vector field and can be computed trivially for samples $x^0, \, \varepsilon$. The power of flow matching lies in learning $v_\theta$ by regressing to $u_\tau(x^\tau \mid x^0)$:
\begin{equation}
\label{eq:objective}
    \mathcal{L}_{\text{CFM}} = \mathbb{E}_{\mathcal{T}(\tau),\,p_0(x^0),\,p_\tau(x^\tau \mid x^0)} \left\lVert v_\theta(x^\tau, \tau) - u_\tau(x^\tau \mid x^0) \right\rVert^2,
\end{equation}
where the expectation is taken over a distribution $\mathcal{T}$ of flow times $\tau$, which is $\mathcal{U}([0, 1])$ in \cite{lipman_flow_2023} and will take different values in this work.

Inference is performed by integrating the learned field $v_\theta$ from $\tau=1$ to $\tau=0$ to recover $\hat{x}^0 \sim p_0$:
\begin{equation}
\label{eq:x_hat}
    \hat{x}^0 = \varepsilon + \int_{1}^{0} v_\theta(\hat{x}^\tau, \tau) d\tau
\end{equation}
Critically, this continuous time parameter $\tau$ allows us to define \textit{partial denoising} (stopping at intermediate $\tau > 0$), which is central to our method.

\subsection{Architecture Formulation}
Formally, we aim to learn a generalist robot policy $\pi(\mathbf{A}_t \mid \mathbf{o}_t, \, l)$ that predicts a sequence of actions $\mathbf{A}_t = [\mathbf{a}_t,\dots,\mathbf{a}_{t+H_a-1}]$ given observations consisting of multiple RGB images $\mathbf{I}_{t'}$, a language instruction $l$ and the robot's proprioceptive state $\mathbf{q}_t$, such that $\mathbf{o}_t = [ \mathbf{I}_{t-H_o+1}, \dots, \mathbf{I}_t, l, \mathbf{q}_t ]$.

Our model consists of two flow matching-based models trained using the objective defined in Eq.~\ref{eq:objective}. Let $\mathbf{z}^0_t$ be the sequence of video encodings and $\mathbf{A}^0_t$ be the clean action chunk:

\noindent   \textbf{Video Model:} $v_\phi(\mathbf{z}^0_\text{past}, \, \mathbf{z}^{\tau_v}_\text{future}, \, l, \, \tau_v)$ induces $p_\phi(\mathbf{z}^0_\text{future} \mid \mathbf{z}^0_\text{past}, \, l)$.\\
\noindent \textbf{Action Policy:} $\pi_\theta(\mathbf{A}^{\tau_a}_t, \, \mathbf{q}_t, \, \mathbf{h}^{\tau_v}, \, \tau_a, \, \tau_v)$ induces the action distribution $p_\theta(\mathbf{A}^0_t \mid \mathbf{q}_t, \, \mathbf{h}^{\tau_v}_t, \, \tau_v)$.

Here, $\mathbf{h}^{\tau_v} = v_\phi^{(k)}(\mathbf{z}^0_\text{past}, \, \mathbf{z}^{\tau_v}_\text{future}, \, l, \, \tau_v)$ is the vector of hidden states extracted after the $k^\text{th}$ layer of the video model when invoking it on ``noisy'' video input $\mathbf{z}^{\tau_v}_\text{future}$ (computed via Eq.~\ref{eq:x_tau}) at flow-time $\tau_v$. We illustrate our architecture in Fig. \ref{fig:arch}.

\subsection{Video Model}
While our Video-Action Model formulation can be instantiated with any flow matching-based video model, in practice we use Cosmos-Predict2 \cite{nvidia_world_2025,nvidia2025cosmosworldfoundationmodel} as our base model. Cosmos-Predict2 is an open-source 2B latent Diffusion Transformer (DiT) \cite{peebles2023scalablediffusionmodelstransformers} model that operates on a sequence of video frames encoded by a pretrained 3D-tokenizer. 
The input to the model is a concatenation of clean latent patch embeddings from a context prefix (for which we choose to use 5 frames) and ``noisy'' latent patches representing the future frames to be generated. Each transformer layer alternates between (1) self-attention over the full video sequence, (2) cross-attention to language instructions encoded by T5~\cite{raffel2023exploringlimitstransferlearning}, and (3) a two-layer MLP. %

\subsection{Action Decoder}
The action decoder is instantiated as a DiT that encodes the robot's proprioceptive state $\mathbf{q}_t$ and a sequence of $\mathbf{A}_t$ future robot actions through two separate MLP networks and concatenates them to form the action decoder's sequence dimension. We use learned absolute positional encodings to add temporal information to each token. During training, we randomly replace the soft token encoding the proprioceptive state with a learned mask token to prevent overfitting on the low dimensional observation. 
Each action decoder layer consists of (1) cross-attention to intermediate video model representations $\mathbf{h}^{\tau_v}$, (2) self-attention over the action sequence, and (3) a two-layer MLP. Each component is bypassed by a residual path and each component's output is modulated via AdaLN \cite{peebles2023scalablediffusionmodelstransformers}, where the input to the AdaLN projections is a low-rank bilinear-affine encoding of both video and action flow times $\tau_v$ and $\tau_a$.

\subsection{Action Sampling}
To enable real-time control, we formulate inference as efficient sampling from the marginal action policy. Although \model is in principle capable of sampling from the joint video-action distribution (see Fig. \ref{fig:embodiments} for an example), we can sample from the marginal action distribution more efficiently by bypassing the computational cost of full video reconstruction. We therefore propose a \textit{partial denoising} strategy that extracts semantic features from intermediate flow states without resolving fine-grained pixel details.
Our inference action sampling procedure is described in Algorithm \ref{alg:action_sampling}. Given image observations $\mathbf{o}_t$, we integrate the video flow field from Gaussian noise to an intermediate flow time
 $\tau_v$ (see Eq.~\ref{eq:x_hat}). This yields a partially denoised latent state $\mathbf{z}^{\tau_v}_\text{future}$ that retains sufficient structural information to guide the policy.
 We process this state with the first $k$ layers of the video model and pass the resulting activations as conditioning information to the action decoder. The action decoder then performs a full denoising procedure to produce a chunk of robot actions $\mathbf{A}^0_t$.

\begin{algorithm}
\caption{Action Sampling($k, \, \tau_v$)}
\label{alg:action_sampling}
\begin{algorithmic}[1]
\setlength{\itemsep}{2pt}
\State \textbf{Input:} $\mathbf{z}^0_{\text{past}}, \, \mathbf{q}_t,\, l$
\State $\mathbf{z}^1_{\text{future}}, \, \mathbf{A}^1_t \; \sim \; \mathcal{N}(\mathbf{0}, \mathbf{I})$
\vspace{5pt}
\State $\mathbf{z}^{\tau_v}_\text{future} \gets \mathbf{z}^1_{\text{future}} + \int_1^{\tau_v} v_\phi(\mathbf{z}^0_\text{past}, \, \mathbf{z}^{\tau_v'}_\text{future}, \, l, \, \tau_v')\, d\tau_v'$
\State $\mathbf{h}^{\tau_v} \gets v_\phi^{(k)}(\mathbf{z}^0_\text{past}, \, \mathbf{z}^{\tau_v}_\text{future}, \, l, \, \tau_v)$
\vspace{4pt}
\State $\mathbf{A}^0_t \gets \mathbf{A}^1_t + \int_1^0 \pi_\theta(\mathbf{A}^{\tau_a}_t, \, \mathbf{q}_t, \, \mathbf{h}^{\tau_v}_t, \, \tau_a, \, \tau_v) d\tau_a$
\State \Return $\mathbf{A}^0_t$
\end{algorithmic}
\end{algorithm}

At inference time, $\tau_v$ is a free hyperparameter. Its optimal value is task-dependent, but we show in Sec.~\ref{sec:tau_v_results} empirically that it is generally close to $1$ (high noise). In the special case of $\tau_v = 1$, a single forward pass of the computationally intensive video backbone is sufficient to generate a chunk of actions (line 3 in Algorithm~\ref{alg:action_sampling} becomes redundant), facilitating real-time inference in our experiments. We find that $\tau_v = 1$ is a good default value that balances policy performance and inference speed. See Sec.~\ref{sec:tau_v_discussion} for a discussion on the motivation behind ``noisy'' video conditioning.

\subsection{Training}

Video-Action Model training proceeds in two distinct phases operating on disjoint sets of parameters.
The first stage focuses on the video backbone. To align the generalist backbone with the specific visual domain and dynamics of our robotic tasks, we finetune it using Low-Rank Adapters (LoRA)~\cite{hu2021loralowrankadaptationlarge} on robotics video datasets. This adaptation step ensures the model captures domain-specific semantics while preserving its pretrained temporal reasoning capabilities.

The second stage focuses on learning the action decoder $\pi_\theta$ while keeping the video backbone frozen. We train the decoder from scratch to regress the action flow field, conditioned on video representations $\mathbf{h}^{\tau_v}$ extracted from the frozen backbone. Crucially, to ensure robustness to varying noise levels during inference, we sample \textit{independent} flow times $\tau_v$ (for video) and $\tau_a$ (for action) during each training iteration, as detailed in Algorithm~\ref{alg:action_decoder_training}. We employ a logit-normal distribution for $\mathcal{T}_v$ matching the video pretraining and $\mathcal{T}_a(\tau_a) \propto \sqrt{\tau_a - 0.001}$ for actions following~\cite{black_pi0_2024}. This decoupled training scheme renders our approach significantly more sample-efficient and faster to converge than comparable VLA baselines (see Sec.~\ref{sec:results_dataefficiency}).
\begin{algorithm}
\caption{Action Decoder Training($k, \, \mathcal{T}_v, \, \mathcal{T}_a$)}
\label{alg:action_decoder_training}
\begin{algorithmic}[1]
\setlength{\itemsep}{2pt}
\Repeat
    \State $\mathbf{z}_0^{\text{past}},\, \mathbf{z}_0^{\text{future}}, \, \mathbf{a}_0,\, \mathbf{s}_0,\, l \; \sim \; p_0(\mathbf{z}_0^{\text{past}},\, \mathbf{z}_0^{\text{future}},\, \mathbf{a}_0,\, \mathbf{s}_0,\, l)$
    \State $\tau_v \sim \mathcal{T}_v(\tau_v); \enspace \tau_a \sim \mathcal{T}_a(\tau_a)$
    \State $\boldsymbol{\varepsilon}_v,\, \boldsymbol{\varepsilon}_a \sim \mathcal{N}(\mathbf{0}, \mathbf{I})$
    \vspace{4pt}
    \State $\mathbf{z}_{\tau_v}^\text{future} \gets (1-\tau_v)\, \mathbf{z}_0^\text{future} + \tau_v \, \boldsymbol{\varepsilon}_v$
    \State $\mathbf{a}_{\tau_a} \gets (1-\tau_a)\, \mathbf{a}_0 + \tau_a \, \boldsymbol{\varepsilon}_a$
    \vspace{4pt}
    \State $\mathbf{h}_{\tau_v} \gets v_\phi^{(k)}(\mathbf{z}_0^\text{past}, \, \mathbf{z}_{\tau_v}^\text{future}, \, l, \, \tau_v)$
    \State Take gradient descent step on
    \Statex \quad $\nabla_\theta \left\| \pi_\theta(\mathbf{a}_{\tau_a}, \, \mathbf{s}_0, \, \mathbf{h}_{\tau_v}, \, \tau_a, \, \tau_v) - u_{\tau_a}(\mathbf{a}_{\tau_a} \mid \mathbf{a}_0) \right\|^2$
\Until{converged}
\end{algorithmic}
\end{algorithm}

\section{Experiments}
\label{sec:exp}

\begin{figure*}[t]
    \centering
    \includegraphics[width=0.89\textwidth]{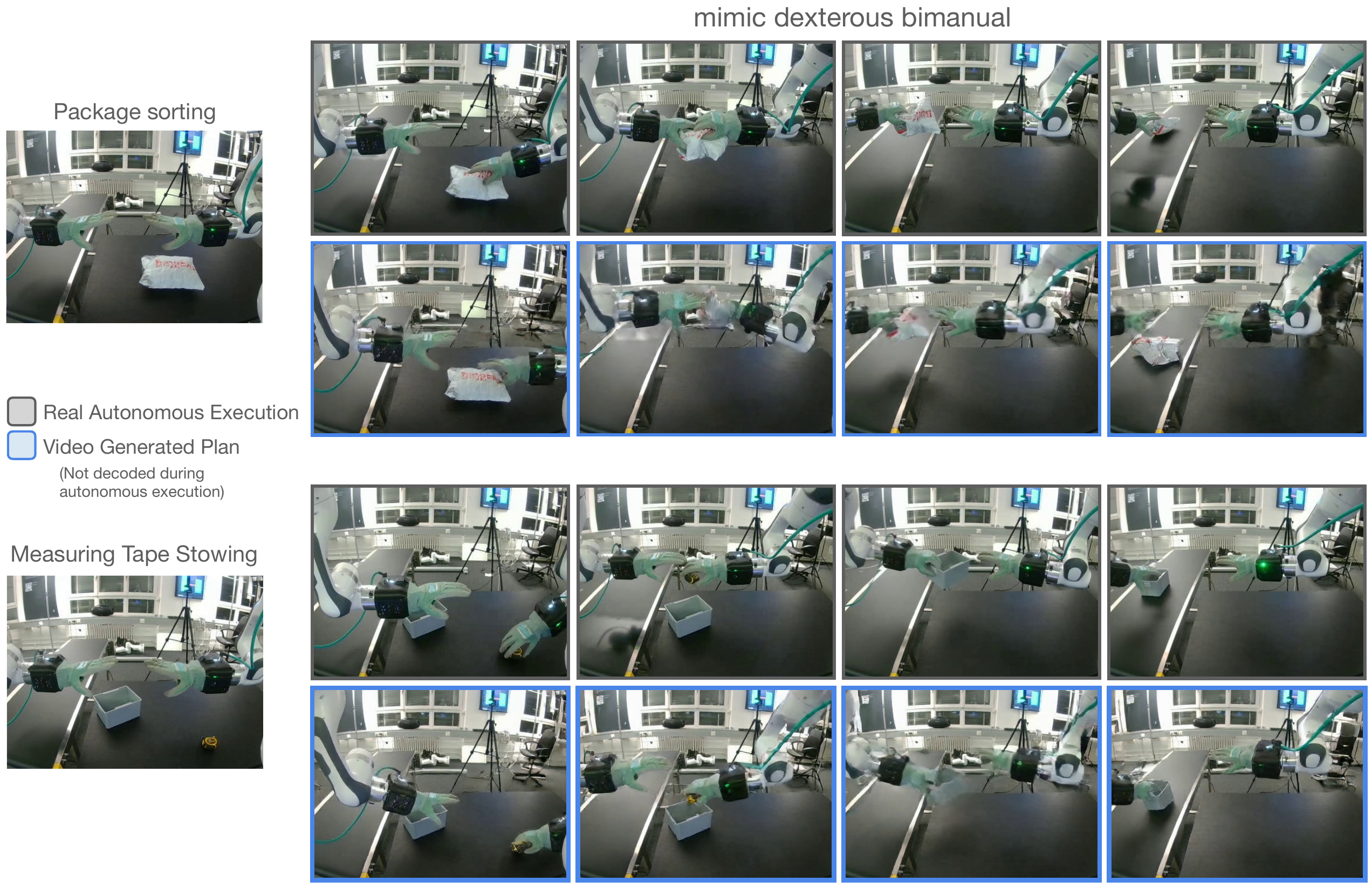}
    \caption{We train and evaluate \model on a real-world bimanual robot setup with Franka Emika Panda robot arms and mimic 16-DoF dexterous humanoid hands. We execute a real-world evaluation on the bimanual setup with two tasks: package sorting and pick and place of a measuring tape into a box. For each action chunk, mimic-video generates a latent video plan ($\tau_v = 1$) and then executes the actions on the real robot. We further fully denoise the predicted video for this visualization.}
    \label{fig:embodiments}
    \vspace{-0.5cm}
\end{figure*}

Our experiments provide an empirical analysis of \model, evaluating the efficacy of leveraging a video backbone for robotic control across several axes:

\begin{enumerate}
\item Can \model effectively control multiple embodiments?
\item Does conditioning on a generative video backbone yield superior sample efficiency and faster convergence for action decoder training, compared to conditioning on a VLM backbone?
\item Is fine-grained video reconstruction necessary for effective policy learning?
\end{enumerate}

\paragraph{Evaluation setups}
We evaluate \model's capabilities across the simulated benchmarks SIMPLER \cite{li2024evaluatingrealworldrobotmanipulation} and LIBERO \cite{liu2023liberobenchmarkingknowledgetransfer}, as well as through real-world dexterous manipulation experiments using humanoid hands.

\textbf{SIMPLER} serves as a high-fidelity proxy for real-world performance, evaluating policies trained on the BridgeDataV2 \cite{walke2024bridgedatav2datasetrobot} dataset of a Widow-X robot embodiment. By employing  system identification and visual matching, it specifically tests the policy's ability to generalize to unseen tasks under realistic visual domain shifts.

\textbf{LIBERO} benchmark evaluates precision and multi-task capacity with a simulated tabletop Panda robot. We focus on the LIBERO-Goal, -Object, and -Spatial suites, each providing 50 expert demonstrations per task across 10 distinct tasks. This widely used benchmark assesses the model's ability to learn precise multi-task manipulation behaviors.

\textbf{Real-World Dexterous Bimanual Manipulation}. To validate \model on high-dimensional, contact-rich tasks, 
we utilize a bimanual setup equipped with two 16-DoF ``mimic'' hands~\cite{nava_mimic-one_2025} mounted on Panda arms. The observation space includes a global workspace view, four wrist cameras, and full proprioception. We evaluate on two long-horizon tasks: Package Sorting (pick, handover, place) and Tape Stowing (pick, stow, move box). Critically, while the video backbone is finetuned on a broader 200-hour corpus, the respective action decoders are trained on extremely scarce task-specific data: just 1h 33m (512 episodes) for sorting and 2h 14m (480 episodes) for stowing.

\paragraph{Comparisons}
\label{paragraph:comparisons}
We compare \model's ability to control multiple robots  against several state-of-the-art baselines.

\textbf{$\pi_{0.5}$-style VLA (Knowledge-Insulating).} To isolate the effect of video pretraining versus standard vision-language pretraining, we construct a VLM-based baseline following a similar architecture as $\pi_{0.5}$~\cite{intelligence__05_2025,driess2025knowledge}. We employ the 3B-parameter PaliGemma~\cite{beyer2024paligemmaversatile3bvlm} as backbone, coupled with an action decoder identical to that of \model. Mirroring \model, the action decoder cross-attends to a particular layer of the backbone for which we empirically find the optimal choice. This equivalent action decoder design, together with training on perfectly equivalent datasets, ensures that performance differences in our comparisons stem strictly from the quality of the conditioning representations (video vs. image-text).
Adhering to the ``Knowledge-Insulation'' protocol~\cite{driess2025knowledge}, we employ a two-stage training process: the autoregressive backbone is trained via Next Token Prediction (discretizing and compressing actions with FAST~\cite{pertsch25-fast}), while the action decoder is separately trained via flow matching.
Notably, while the original $\pi_{0.5}$ leverages massive web and cross-embodiment pretraining, our baseline (denoted ``$\pi_{0.5}$-style'') trains the backbone from the original VLM checkpoint and the decoder from scratch. This standardizes the data regime across methods, allowing for a fair evaluation of the backbone prior.

\begin{table*}[ht]
\caption{
Benchmark scores on SIMPLER-Bridge. The training regimes denote the usage of robot action data: ``pretrained'' (large-scale external), ``finetuned'' (external $\to$ target), and ``scratch'' (target only). Note that all models leverage image or video pretraining. \textbf{Bold}: best overall; \underline{underline}: best ``scratch'' score. We also report \model with task-optimized $\tau_v$.}

\label{tab:results-two-finger-simplerenv}
\centering
\renewcommand{\arraystretch}{1.15}
\setlength{\tabcolsep}{10pt}
\begin{tabular}{lccccc}
\hline
\multicolumn{6}{c}{\begin{tabular}[c]{@{}c@{}}
Inputs: third-person image, language instruction, robot proprioceptive state (optional)
\end{tabular}} \\
\hline
Model & Put Carrot on Plate & Put Spoon on Towel & Stack Blocks & Eggplant & Average SR (\%) \\
\hline
OpenVLA (finetuned) \cite{kim_openvla_2024} & 4.2 & 8.3 & 0.0 & 45.8 & 14.6 \\
Octo (finetuned) \cite{team_octo_2024} & 8.3 & 12.5 & 0.0 & 43.1 & 16.0 \\
ThinkAct (pretrained) \cite{huang2025thinkactvisionlanguageactionreasoningreinforced} & \textbf{37.5} & 58.3 & 8.7 & 70.8 & 43.8 \\
FLOWER (finetuned) \cite{reuss2025flowerdemocratizinggeneralistrobot} & 13.0 & \textbf{71.0} & 8.0 & 88.0 & 45.0 \\
\hline
$\pi_{0.5}$-style VLA (scratch) & 25.0 & 29.2 & \textbf{\underline{20.8}} & 66.7 & 35.4 \\
\textbf{\model} (scratch) & \textbf{\underline{37.5}} & \underline{37.5} & 12.5 & \textbf{\underline{100.0}} & \textbf{\underline{46.9}} \\
\hline
\textbf{\model} (scratch, per task $\tau_v$-tuning) & 54.2 & 41.7 & 29.2 & 100.0 & 56.3 \\
\hline
\end{tabular}
\end{table*}

\textbf{DiT-Block Policy.} For real-world bimanual evaluations, we compare against a strong single-task baseline: a DiT-Block Policy~\cite{dasari2024ingredients} following the action representation recipe from \citet{nava_mimic-one_2025}. This model features a ViT-S DINO backbone~\cite{dosovitskiy_image_2021, caron_emerging_2021} (with separate encoders for each camera view) feeding into an 8-block, 8-head transformer diffusion policy. With approximately 155M parameters (in the multi-view setting), this architecture represents a competitive standard for imitation learning in low-data regimes, making it an ideal reference point for the utilized bimanual dexterous tele-operation datasets.

\textbf{State-of-the-Art Published Baselines.} We additionally include results reported for state-of-the-art competing approaches, namely Octo \cite{team_octo_2024}, ThinkAct \cite{huang2025thinkactvisionlanguageactionreasoningreinforced}, FLOWER \cite{reuss2025flowerdemocratizinggeneralistrobot}, OpenVLA \cite{kim_openvla_2024} and OpenVLA-OFT \cite{kim2025finetuningvisionlanguageactionmodelsoptimizing}.

\subsection{Direct Evaluation across Diverse Robot Platforms}
\paragraph{SIMPLER-Bridge}
We first evaluate \model's cross-task generalization capabilities on the SIMPLER-Bridge benchmark, with full results detailed in Table~\ref{tab:results-two-finger-simplerenv}. Our model achieves the strongest average success rate across all four tasks, matching or surpassing the performance of state-of-the-art baselines, including our $\pi_{0.5}$-style VLA comparison. This strong performance validates that conditioning on the generative video prior yields more robust policy representations than those derived from vision-language-action (VLA) pretraining alone. Additionally, leveraging the partial denoising strategy, we demonstrate a novel form of \textit{inference-time policy optimization}: by adjusting the flow parameter $\tau_v$, the fixed trained model can be specialized to individual task dynamics, achieving further performance gains at the cost of modest increases in computation.

\begin{table}[h]
\caption{Benchmark scores on LIBERO. ``finetuned'', ``scratch'', \textbf{bold}, and \underline{underline} are defined as in Tab. \ref{tab:results-two-finger-simplerenv}.}
\label{tab:results-libero}
\begin{center}
\setlength{\tabcolsep}{1.5pt}
\begin{tabular}{lcccc}
\hline
\multicolumn{5}{c}{\begin{tabular}[c]{@{}c@{}}
 Inputs: third-person image, language instruction, proprioception (optional) \\
\end{tabular}} \\
\hline
Model & Spatial  (\%) & Object  (\%) & Goal  (\%) & Avg (\%) \\
\hline
Diffusion Policy (scratch) \cite{chi_diffusion_2024} & 78.3 & 92.5 & 68.3 & 79.7 \\
Octo (finetuned) \cite{team_octo_2024} & 78.9 & 85.7 & 84.6 & 83.1 \\
DiT Policy (finetuned) \cite{dasari2024ingredients} & 84.2 & 96.3 & 85.4 & 88.6 \\
OpenVLA (finetuned) \cite{kim_openvla_2024} & 84.7 & 88.4 & 79.2 & 84.1 \\
OpenVLA-OFT (finetuned) \cite{kim2025finetuningvisionlanguageactionmodelsoptimizing} & \textbf{96.2} & \textbf{98.3} & \textbf{96.2} & \textbf{96.9} \\
\hline
$\pi_{0.5}$-style VLA (scratch) & 79.2 & 94.0 & 84.4 & 85.9 \\
\textbf{\model} (scratch) & \underline{94.2} & \underline{96.8} & \underline{90.6} & \underline{93.9} \\
\hline
\end{tabular}
\end{center}
\end{table}

\paragraph{LIBERO}
We evaluate \model's multi-task manipulation capabilities on the LIBERO benchmark. Despite being trained from scratch on task-specific action data, \model outperforms the majority of state-of-the-art methods finetuned from generalist models (see Table~\ref{tab:results-libero}). Notably, \model achieves significantly higher success rates than the comparable $\pi_{0.5}$-style VLA baseline, indicating that the generative video prior facilitates more robust and efficient policy learning than the corresponding vision-language pretrained representations.

\paragraph{Real-World Dexterous Bimanual System}
To validate \model on high-dimensional, contact-rich tasks under real-world data scarcity, we benchmark it against single-task DiT-Block Policies on a bimanual setup comprising two Franka arms equipped with dexterous humanoid hands. This setup presents a significant challenge due to heavy occlusions, particularly during grasping, where wrist camera observations play a critical role in guiding  robot policies. This necessity is reflected by the performance gap between the two DiT-Block Policy variants (workspace-only vs. multi-view) shown in Table~\ref{tab:results-dexterous}. Remarkably, \model significantly surpasses the performance of both baselines, despite only being conditioned on the single workspace camera view. This result confirms that the predictive capacity of the generative video prior allows \model to effectively bridge the visual uncertainty caused by occlusion, leading to robust policies learned from minimal task-specific data. Fig.~\ref{fig:embodiments} illustrates the real world experiment.

\begin{table}[h]
\caption{Benchmark scores on real-world bimanual dexterous manipulation on the mimic system.}
\label{tab:results-dexterous}
\begin{center}
\setlength{\tabcolsep}{3pt}
\begin{tabular}{lcc}
\hline
Model & Packing & Package handover\\
\hline
 DiT-Block Policy~\cite{dasari2024ingredients}   & 11.0 & 30.0 \\
 DiT-Block Policy~\cite{dasari2024ingredients}  (+ wrist cams) & 42.6 & 74.1\\
\textbf{\model}  & \textbf{72.0} & \textbf{93.0} \\
\hline
\end{tabular}
\end{center}

\end{table}

\begin{figure}[b]
    \centering
    \includegraphics[width=0.35\textwidth]{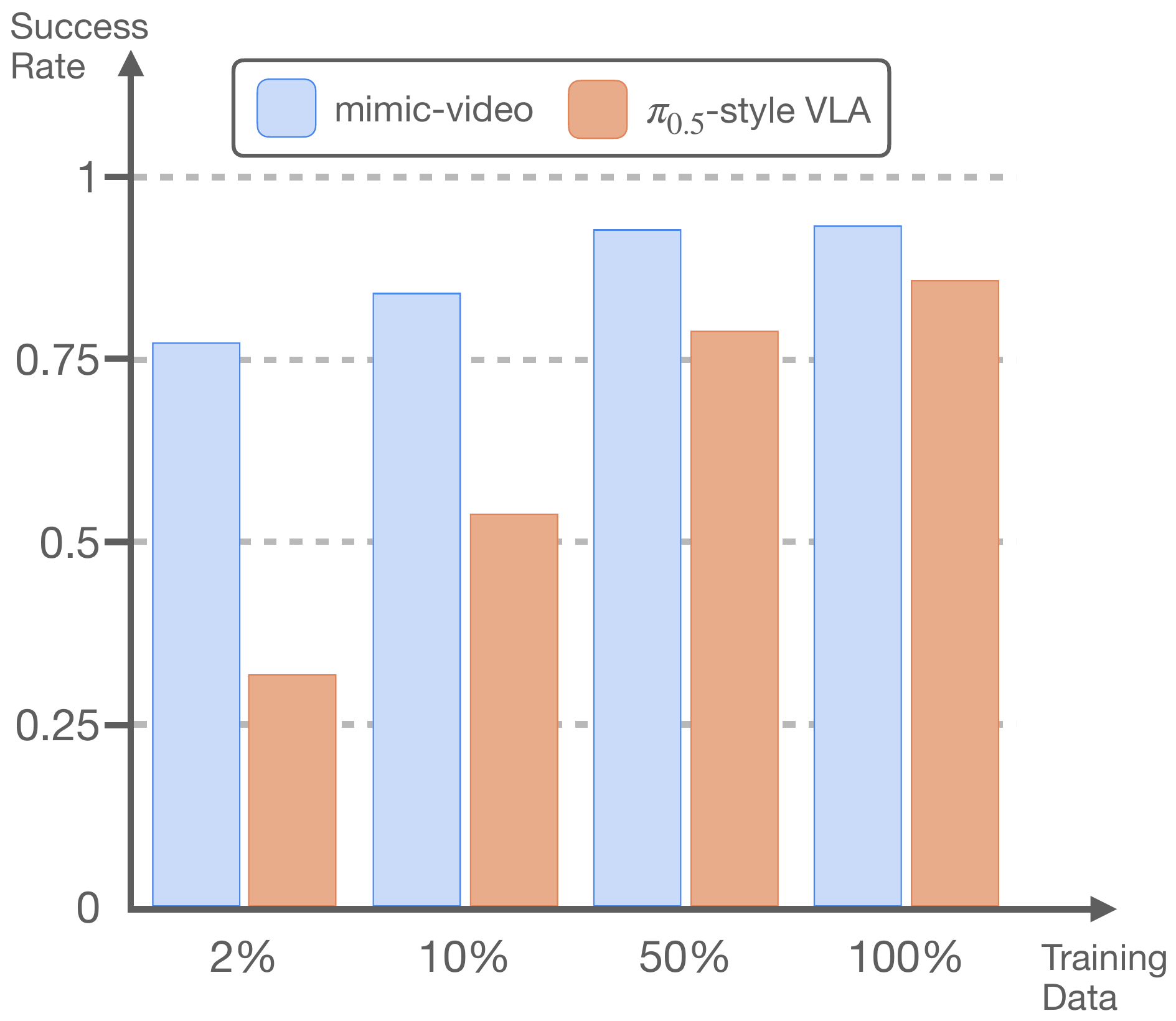}
    \caption{Sample efficiency for action decoder training on LIBERO: \model against the $\pi_{0.5}$-style VLA baseline.}
    \label{fig:sample_efficiency}
\end{figure}

\subsection{Data Efficiency and Convergence Speed}
\label{sec:results_dataefficiency}
We investigate the data efficiency of decoding actions from video model representations compared to the VLM representations by training \model and $\pi_{0.5}$-style VLA action decoders on differently-sized subsets of the LIBERO-Goal, LIBERO-Spatial, and LIBERO-Object task suites. The result, shown in Fig. \ref{fig:sample_efficiency}, demonstrates a remarkable \textit{order-of-magnitude increase in sample efficiency}
when conditioning on the video prior. Specifically, \model's action decoder reaches the maximum success rate achieved by the VLM-conditioned decoder while requiring only  10\% of the training data. Decreasing the dataset size to only one episode per task (a $98\%$ reduction in action data), still yields a 77\% average success rate, placing \model trained on 2\% of the action data competitive with our Diffusion Policy baseline.

\begin{figure}[t]
    \centering
    \includegraphics[width=0.42\textwidth]{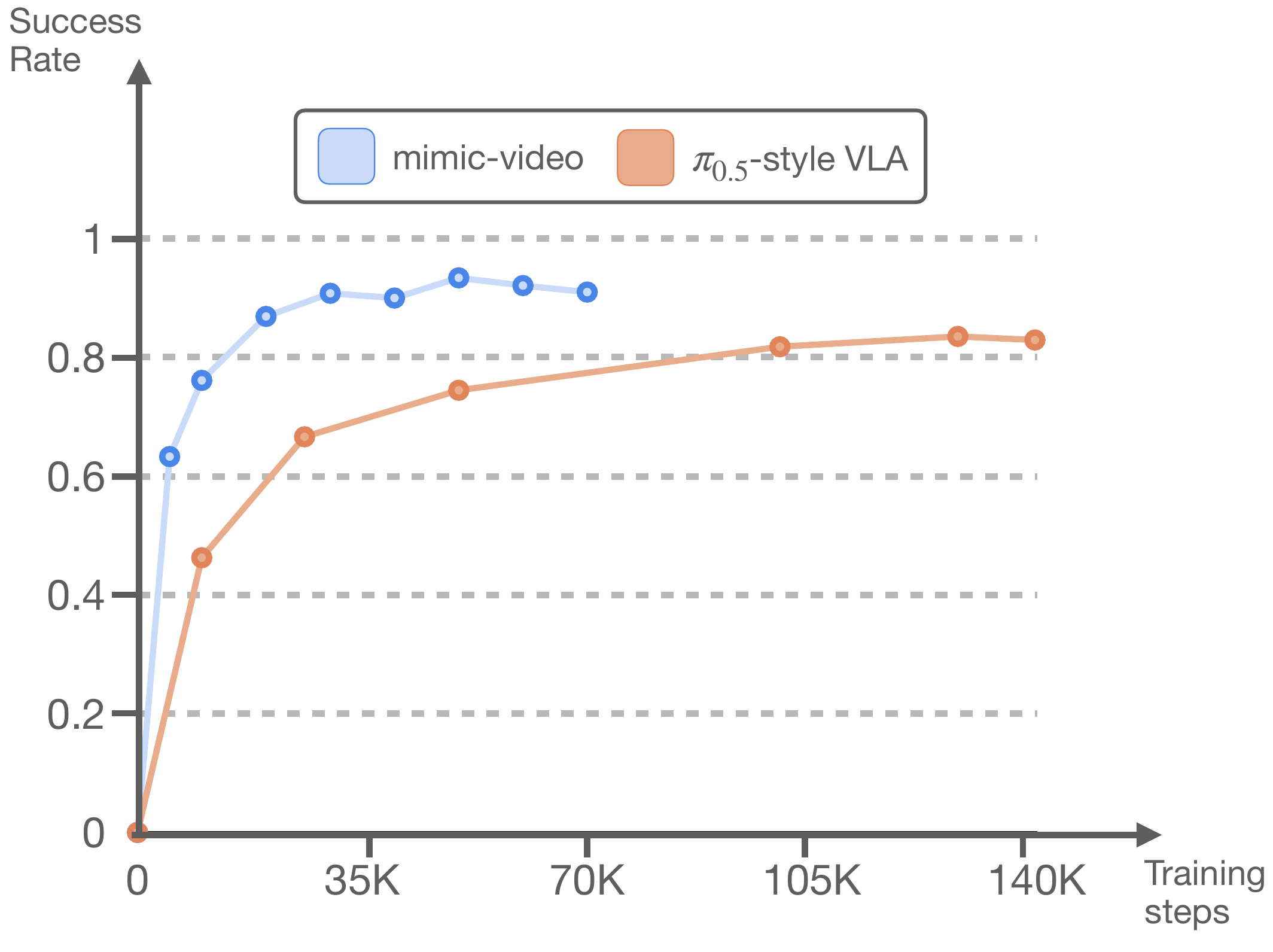}
    \caption{Convergence speed for action decoder training. Both decoders are trained with a batch size of 128 (optimal for $\pi_{0.5}$-style VLA) and their respective optimal learning rate.}
    \label{fig:convergence}
\end{figure}

Beyond sample efficiency, Fig.~\ref{fig:convergence} shows that the \model action decoder converges significantly faster and to a higher asymptotic success rate than the $\pi_{0.5}$-style VLA decoder. Notably, this advantage persists despite the VLA baseline having been exposed to task-specific action data during FAST-pretraining. 

\subsection{Trade-offs between Video Fidelity and Action Performance}
\label{sec:tau_v_results}

\begin{figure}[t]
    \centering
    \includegraphics[width=0.4\textwidth]{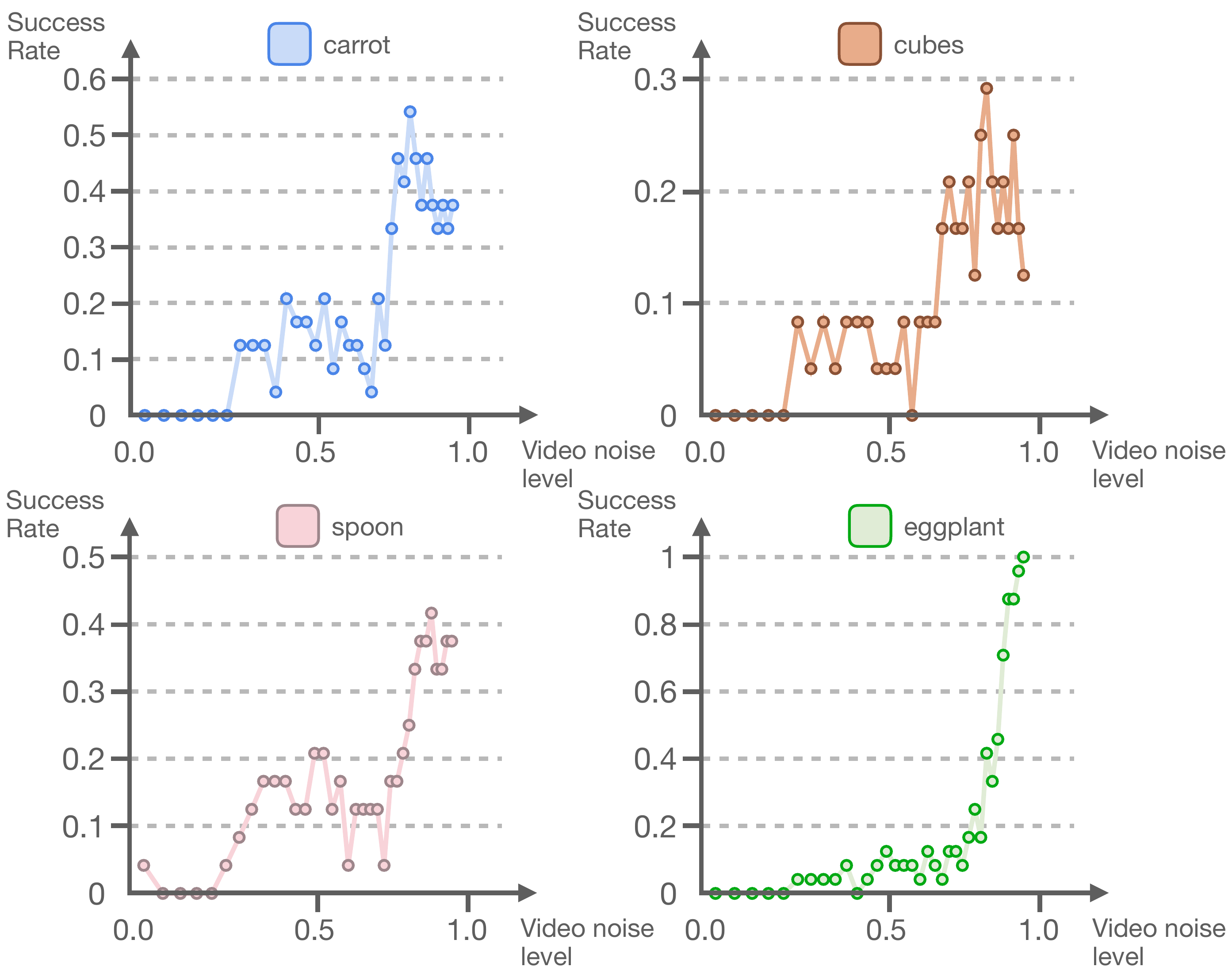}
    \caption{Policy success rate across the SIMPLER-Bridge environments vs video flow time ($\tau_v$, logit-scaled). Performance peaks at an intermediate noise level, confirming that high-fidelity video reconstruction is not required for performant robot policies.}
    \label{fig:optimal-noise}
\end{figure}

\model couples two separate flow matching models for video and actions, respectively. A key design element of our approach is the ability to control the video generation process via an inference-time hyperparameter: the video flow time $\tau_v \in [0, 1]$. This parameter dictates the extent to which future video latents are denoised during action sampling. To investigate the necessity of fine-grained video reconstruction for effective policy learning, we first note the intuitive hypothesis: a more resolved, higher-fidelity video signal should correlate with better policy performance. In order to study this question, we sweep $\tau_v$ across the SIMPLER-Bridge environments and visualize the resulting success rates in Fig. \ref{fig:optimal-noise}.

\begin{figure}[b]
    \centering
\includegraphics[width=0.4\textwidth]{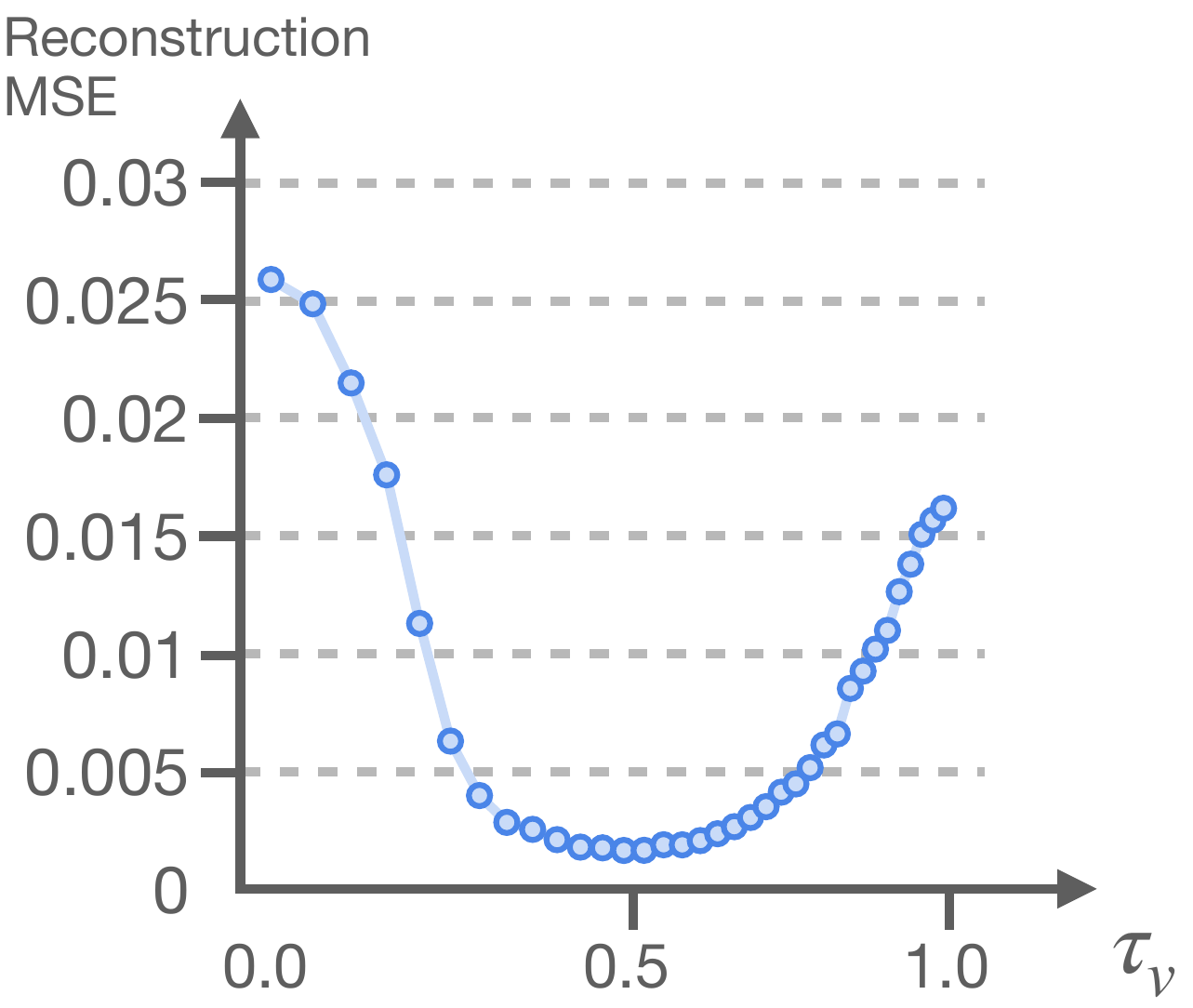}
    \caption{Action reconstruction MSE of a decoder conditioned on ``noisy'' ground-truth video latents at varying flow times (logit-scaled) on BridgeDataV2. Reconstruction is best at intermediate flow times and increases towards  clean and pure noise latents.}
    \label{fig:mse}
\end{figure}

Counterintuitively, we find that the best autonomous policy performance in our SIMPLER experiments is achieved at the highest flow time $\tau_v = 1$.
Theoretically, as $\tau_v$ progresses from $1$ (pure noise) to $0$ (full reconstruction), the underlying video signal grows, and the mutual information $I(\mathbf{z}^{\tau_v}_\text{future} ; \mathbf{A}^0)$ between the future video latent and future actions increases. 
However, consistent with the case study in Sec. \ref{sec:case-study}, we hypothesize that imperfect video generation introduces artifacts.
Consequently, fully denoised video latents may diverge from the training distribution, presenting out-of-distribution conditioning to the action decoder. To isolate the effect of these generation errors, we perform an additional sweep of $\tau_v$ where we condition the action decoder on ``noisy'' ground-truth video latents, $\mathbf{z}_{\tau_v}^\text{future}$, computed via Eq. \ref{eq:x_tau}. We report the resulting action reconstruction MSE on a held-out validation set of BridgeDataV2 in Fig. \ref{fig:mse}.
We observe that the lowest action reconstruction error is achieved at an intermediate flow time of $\tau_v \approx 0.4$, corresponding to the perfect rollout performance observed in our  Case Study (Sec. \ref{sec:case-study}).

Interestingly, action prediction error increases sharply as we move from this optimum towards $\tau_v = 0$ (full reconstruction). We attribute this to the nature of the conditioning signal: while the \textit{video latents} themselves contain more information at lower noise, the \textit{intermediate video model representations}—from which the action decoder reconstructs actions—may exhibit distinct, non-trivial behavior.
We provide a detailed discussion of these mechanisms in Appendix \ref{sec:tau_v_discussion}. 
This observation yields a significant practical advantage: operating at $\tau_v = 1$ requires only a single forward pass of the video backbone to generate conditioning features, resulting in both the highest average performance and the fastest inference speed.

\section{Discussion and Future Work}
\label{sec:discussion}
In this work, we introduce \model, a new class of Video-Action Model (VAM) that grounds robotic policies in a pretrained video model. By leveraging the physical priors embedded in internet-scale video, \model{} achieves an order-of-magnitude improvement in sample efficiency and significantly faster convergence compared to standard VLA baselines. These results strongly suggest that representations learned from large-scale generative video pretraining provide a significantly more robust signal for policy learning than those induced by vision–language-action pretraining.
To achieve this, our approach operates by first partially generating a plausible video of a task's successful execution. We find that conditioning on these partially-denoised plans is critical, yielding a dual benefit: it mitigates the distribution shift between model predictions and the ground-truth data used for training, while simultaneously accelerating inference by significantly reducing the computational cost of video generation.

While \model achieves strong performance across both simulated and real-world evaluations, we find that the current model still has several shortcomings. First, we rely on a single-view video backbone, which restricts our policies to a fixed, single workspace view. Exploring a wider range of video architectures, particularly natively multi-view models,  would likely enhance spatial reasoning and occlusion robustness. Second, we have not yet applied the VAM recipe to train a unified, large-scale, cross-embodiment model, a step we believe is necessary to unlock the full  generalization capabilities of video foundation models. Finally, our current real-world experiments are limited to a focused set of tasks; scaling this approach to a broader diversity of manipulation behaviors remains a key objective for future work.

\section*{Acknowledgments}

This work was supported under project ID \#36 as part of the Swiss AI Initiative, through a grant from the ETH Domain and computational resources provided by the Swiss National Supercomputing Centre (CSCS) under the Alps infrastructure. We thank mimic robotics for providing experimental infrastructure, real-world robot platforms and additional compute resources. Primary work by the lead authors was performed during their internships at mimic robotics, with continued development supported during their internships at Microsoft.

We thank Benjamin Estermann, Stefanos Charalambous, Erik Bauer, German Rodriguez, Sigmund Hennum Høeg, Irvin Totic and Benedict Wüest for their help with the project.

\bibliographystyle{plainnat}
\bibliography{references-new}

\clearpage

\section*{Contributions}

\noindent\textbf{Jonas Pai:} Led project ideation, implementation, and evaluation. Contributed to tech report writing.

\noindent\textbf{Liam Achenbach:} Led baseline model development, training, and evaluation. Helped with dataset integration and tech report writing.

\noindent\textbf{Victoriano Montesinos:} Contributed the diffusion policy baseline implementation and training, as well as data collection for the bimanual mimic robot experiments.

\noindent\textbf{Benedek Forrai:} Oversaw development for the mimic bimanual system, contributed data collection for the bimanual mimic robot experiments.

\noindent\textbf{Oier Mees:} Supervised the project from its inception, mentored the lead authors during their research internships, guided the technical direction and experimental strategy, and led the writing and visualization of this tech report.

\noindent\textbf{Elvis Nava:} Supervised the project from early conception to implementation and evaluations, contributed to project ideation. Supervised the lead authors throughout their research internship and oversaw the technical development of supporting infrastructure, robot systems and compute resources. Contributed to manuscript writing, website and video editing. Contributed data collection for the bimanual mimic robot experiments.

\appendix

\subsection{Training Hyperparameters}
We summarize \model training hyperparameters for each dataset in Tab.~\ref{tab:hyperparameters-vam}.
\begin{table*}[h]
\caption{Hyperparameters used during training of \model.}
\label{tab:hyperparameters-vam}
\centering
\renewcommand{\arraystretch}{1.15}
\setlength{\tabcolsep}{10pt}
\begin{tabular}{l ccc ccc}
\hline
\multirow{2}{*}{\textbf{Hyperparameter}}
& \multicolumn{3}{c}{\textbf{Video finetuning}}
& \multicolumn{3}{c}{\textbf{Action Decoder training}} \\
& \textbf{BridgeDataV2} & \textbf{LIBERO} & \textbf{mimic}
& \textbf{BridgeDataV2} & \textbf{LIBERO} & \textbf{mimic} \\
\hline
Learning Rate           
& \multicolumn{3}{c}{1.778e-4}
& \multicolumn{3}{c}{1e-4} \\

Warmup Steps            
& \multicolumn{6}{c}{1000} \\

Training Steps
& 70043 & 7k-8k & 27300 & 14112 & 50k & 26k \\

LR Scheduler            
& \multicolumn{3}{c}{Constant}
& \multicolumn{3}{c}{Linear} \\

Weight Decay Factor
& \multicolumn{6}{c}{0.1} \\

Gradient Clip Threshold 
& \multicolumn{6}{c}{10.0} \\

Batch Size              
& 256 & 128 & 32
& 256 & 128 & 128 \\

Optimizer
& \multicolumn{6}{c}{AdamW \cite{loshchilov2019decoupledweightdecayregularization}} \\
\hline
\end{tabular}
\end{table*}

\subsection{Data Preprocessing}

All orientations are expressed as 6-dimensional vectors corresponding to the top two rows of the matrix representation. Images are extracted or rendered in a resolution of 480 x 640 px.

\paragraph{BridgeDataV2}

\begin{itemize}
    \item Observation Space: Absolute end-effector pose and absolute continuous gripper joint state.
    \item Action Space: Future end effector pose (relative to the proprioceptive pose for the entire action chunk) and the continuous (but practically mostly binary) gripper \textit{action}.
\end{itemize}
We remove 3046 non-informative language labels as well as the first state and null-action of each episode.

\paragraph{LIBERO}

\begin{itemize}
    \item Observation Space: Absolute end-effector pose and absolute continuous gripper joint state.
    \item Action Space: End effector pose action (relative to the proprioceptive pose for the entire action chunk) and binary gripper action.
\end{itemize}
We follow the preprocessing procedure used in \citet{kim2025finetuningvisionlanguageactionmodelsoptimizing} and remove episodes not leading to a successful rollout when replaying their actions.

\paragraph{mimic}

\begin{itemize}
    \item Observation Space: Absolute end-effector poses, absolute continuous hand joint states. Relative end-effector poses with respect to each other. Previous end-effector and hand actions.
    \item Action Space: End effector pose action (relative to the proprioceptive pose for the entire action chunk) and absolute hand joints action.
\end{itemize}

\subsection{Video-Action Model Learnings}
\begin{itemize}
    \item Video Model Source Layer $k$: We observe intermediate layer $k=19$ to yield the strongest policy performance with strongly decreasing success rates towards initial or final layers. We posit that ideally, this choice should be learned.
    \item Video Observation Horizon $H_o$: We find the longer horizon of 5 frames to work better than conditioning on only the current observation (1 frame).
\end{itemize}

\subsection{$\pi_{0.5}$-style VLA Learnings}
We ablate various choices in the VLA training procedure and architecture to enable a fair comparison to \model. 
\begin{itemize}
    \item \textbf{Architectural details}: We find the highest SIMPLER-Bridge success rates when cross-attending to layer 11 of the FAST-pretrained VLM.
    \item \textbf{Training results}: For SIMPLER-Bridge we find that training longer does not improve success rates significantly after 2-3 epochs of decoder training on a frozen FAST-backbone trained to convergence. For LIBERO, we observe that continuing FAST pretraining slightly beyond the convergence point yields modest downstream gains during the subsequent decoder training stage.
\end{itemize}

\subsection{Video Denoising Analysis}
\label{sec:tau_v_discussion}
A key finding of our work is that the choice of the video model's cutoff flow time, $\tau_v$, is a critical inference hyperparameter. We empirically observe that stopping the video generation process early and conditioning the action decoder on a ``noisy'' visual plan yields substantially better performance than allowing the video model to fully denoise its prediction. We find two main reasons for this phenomenon:

\paragraph{Distribution Mismatch and Noise as Augmentation} The action decoder is trained by conditioning on the video model's representations of \textit{ground truth} future video. A fully denoised video generated by the video model at inference time could represent an incorrect action plan due to the video model's weakness. Even if accurate, it will still likely be subtly out-of-distribution compared to the ground truth data seen during training. By intentionally leaving noise in the visual plan, we perform a kind of train and test-time augmentation. This prevents the action decoder from relying on spurious ground truth visual cues that may not be present in the video model's own generations. This is analogous to findings in goal-conditioned policies~\cite{hatch2024videoglue}, where augmenting predicted future target images with simple transformations improved robustness and performance.

The results shown in Fig.~\ref{fig:perf-ours-gt} give credence to this hypothesis, as we observe that conditioning the action decoder on ground truth data at inference time leads to perfect performance, so less-than-perfect performance during regular inference has to be attributed to the fully denoised video plans being imperfect or out of distribution. The results shown in Fig.~\ref{fig:optimal-noise} also illustrate how optimal performance on benchmarks occurs at video flow times close to $\tau_v = 1$, where ``noise as augmentation'' is larger.

\paragraph{Information Content of Intermediate Representations} Another reason lies in the nature of the flow matching model's internal representations throughout the denoising process. In intermediate steps, the hidden states of the video model must encode rich information about scene dynamics and the necessary transformations to reach the final, clean video. However, as the denoising process approaches $\tau_v = 0$, the input is already very close to the target. To minimize the training loss, the video model layers, when conditioned on the final noise values, are incentivized to learn a close-to-identity mapping, making minimal changes to the nearly-perfect input. Consequently, these final-step hidden states become less informative for downstream tasks. Cross-attending to the richer representations from earlier flow times $\tau_v$ provides the action decoder with a more useful conditioning signal for generating actions. The results in Fig.~\ref{fig:mse} indeed distinctly show that, when approaching $\tau_v = 0$, reconstruction error strongly increases, which is compatible with this hypothesis.

\end{document}